\documentclass[conference]{IEEEtran}
\IEEEoverridecommandlockouts
\usepackage{cite}
\usepackage{amsmath,amssymb,amsfonts}
\usepackage{algorithmic}
\usepackage{graphicx}
\usepackage{subfig}
\usepackage{textcomp}
\usepackage{xcolor}
\usepackage{multirow}

\usepackage{eso-pic}

\AddToShipoutPicture*{\small \sffamily\raisebox{1.2cm}{\hspace{1.8cm}978-1-7281-1621-1/19/\$31.00 ©2019 IEEE}}

\def\BibTeX{{\rm B\kern-.05em{\sc i\kern-.025em b}\kern-.08em
    T\kern-.1667em\lower.7ex\hbox{E}\kern-.125emX}}
\begin{document}

\title{Capsule-Based Persian/Arabic Robust Handwritten Digit Recognition Using EM Routing\\
}

\author{\IEEEauthorblockN{Ali Ghofrani}
\IEEEauthorblockA{\textit{Faculty of Media Technology and Engineering} \\
\textit{IRIB University}\\
Tehran, Iran \\
alighofrani@iribu.ac.ir}
\and
\IEEEauthorblockN{Rahil Mahdian Toroghi}
\IEEEauthorblockA{\textit{Faculty of Media Technology and Engineering} \\
\textit{IRIB University}\\
Tehran, Iran \\
mahdian.t.r@gmail.com}

}

\maketitle

\begin{abstract}
In this paper, the problem of handwritten digit recognition has been addressed. However, the underlying language is Persian/Arabic, and the system with which this task is a capsule network (CapsNet) which has recently emerged as a more advanced architecture than its ancestor, namely CNN (Convolutional Neural Network). The training of the architecture is performed using Hoda dataset, which has been provided for Persian/Arabic handwritten digits. The output of the system, clearly outperforms the results achieved by its ancestors, as well as other previously presented recognition algorithms.

\end{abstract}


\section{Introduction}
Handwritten digit recognition is among the very first applications of artificial intelligence in our life, which aimed at automatically reading the zip-codes being written on letters, packages or similar applications. As a branch of image processing, it gained the interest of researchers and many heuristic works were proposed in this regard, \cite{lecun1990handwritten,hochuli2018handwritten,lecun1995comparison,hamidi2010invariance,liu2003handwritten,li2018neural}.

MNIST is the most widely used dataset as a benchmark for handwritten digit recognition tasks. During 70s and 80s, there were lots of MLP-based neural network classifiers among the very first ones being tested on MNIST data.
While this dataset is created for English language, and doing research tasks on such a subject is becoming arguable today, there are only few attempts on Persian or Arabic handwritten digits which are quite different and somehow more difficult to be recognized, in the sense that some of the digits could be written in many different ways \cite{khosravi2007introducing}. 

In this paper, we concentrate on Persian/Arabic-based digital numbers which are the same, however quite different from the English digits gathered in MNIST dataset. Hence, here we use \textbf{Hoda} dataset \cite{khosravi2007introducing}, which is recorded for Persian/Arabic language and alongside a new method being proposed in this work the results are compared with the state-of-the-art \cite{alaei2009fine,salimi2013farsi,sadri2003application,mozaffari2005structural}, even though we strongly believe that this technique could be equally applied to the digits being written in English or other languages, if previously trained.

The novelty of this paper comes in utilizing a capsule network for training and testing the handwritten digit recognition in Persian/Arabic language for the first time.


The outline of this paper, occur as the following: First, the structure of a typical capsule network for the considered purpose is explained. Then, the structure of the system being designed for our digit recognition is mentioned. Section IV, brings about the experimental setup and parameter tunings along with all the requirements for the experiments to be reimplemented. In the sequel, the paper is terminated by the conclusion, and the cited references.

\section{Capsule Networks (CAPSNET)}
While Convolutional Neural Networks (CNNs) happened to be very successful in computer vision applications, they are still weak in understanding the rotation or change of proportion of the input images. CapsNet overcomes these shortcomings and provide local translation invariance (via max-pooling,
typically) by addressing various kinds of visual
stimulus and encoding things such as position, orientation,
deformation, and so on, \cite{sabour2017dynamic,xi2017capsule,xiang2018ms,neill2018siamese}.The complete explanation of capsule networks with Expectation-Maximization (EM) routing mechanism for training of the parameters, are rephrased from \cite{sabour2018matrix,lalonde2018capsules,lin2018learning}.

\subsection{How Capsules Work?}
A capsule network consists of several layers of capsules.
Between two adjacent layers of capsules $L$ and $L+1$, there is a learned transformation matrix $\boldsymbol{W}_{ij}$ for every pair of capsules, $i$ and $j$.
The vote from capsule $i$ to capsule $j$ is computed with a simple matrix multiplication of pose and transformation matrices:
\begin{eqnarray}
\boldsymbol{V}_{ij} = M_i \, \boldsymbol{W}_{ij}
\end{eqnarray}
The pose matrix $M$, represents the relationship between an object or object-part and the pose. In other words, it defines the translation and the rotation of an object which is equivalent to the change of the viewpoint of an object. A capsule in one layer votes for the pose matrix of many different capsules in the layer above, by multiplying its own pose matrix by the transformation matrix $\boldsymbol{W}$, that could learn to represent the part-whole relationship.

The activations $a_i$ and votes $\boldsymbol{V}_{ij}$ for all capsules in layers $i$ and $j$, are used in the routing algorithm to compute the activations
$a_j$ and poses $\boldsymbol{M}_j$ in layer $j$, therefore the routing algorithm decides where the output of the capsule should go. The \textit{dynamic routing by agreement} algorithm uses a routing coefficient that is dependent on the input and gets computed dynamically. Thereby capsule $i$ gets assigned to a capsule $j$ that has the best fitting cluster of votes for its pose.

When given the activations and poses of all capsules in layer $L$ the routing algorithm decides which capsules to activate in layer $L+1$ and how to assign every capsule in layer $L$ to one capsule in layer $L+1$.

In EM routing, we model the pose matrix of the parent capsule with a Gaussian.
The pose matrix is for example a 4×4 matrix, i.e. 16 components. We model the pose matrix with a Gaussian having 16 $\mu$'s and 16 $\sigma$'s, and each $\mu$ represents a pose matrix's component.
Then, we can apply the Gaussian probability density function to compute the
probability of $\boldsymbol{V}_{ij}^h$ (the $h^{\text{th}}$ dimension or component of the vectorized vote) belonging to the capsule $j$'s Gaussian model, as
\begin{eqnarray}
p_{i|j}^h = \frac{1}{\sqrt{2\pi(\sigma_j^h)^2}}\, \text{exp}\bigg(-\frac{(\boldsymbol{V}_{ij}^h-\mu_j^h)^2}{2(\sigma_j^h)^2}  \bigg)
\label{eq:pij}
\end{eqnarray}
\noindent
where $\mu_j$, and $\sigma_j^2$  are the mean and variance of the fitted Gaussian, respectively. We take natural $log$ from (\ref{eq:pij}), and the cost to activate the parent capsule $j$ by the capsule $i$, would be
\begin{eqnarray}
\text{cost}_{ij}^h = - \text{ln} \big( p_{i|j}^h   \big)
\end{eqnarray}
\noindent
Since capsules of a lower layer $i$ are not equally linked with capsule $j$, and 
a capsule with a low activation has no big impact on the routing procedure,
we pro-rate the cost with the runtime assignment probabilities $r_{ij}$, which is called \textit{routing coefficient} between any capsule $i$ and the higher layer capsule $j$. The cost from all lower layer capsules would be then \cite{sabour2018matrix}, 
\begin{eqnarray}
\text{cost}_{j}^h =\sum_{i}r_{ij} \text{cost}_{ij}^h   =\overbrace{\dots}^{Math} = \big(\text{ln}(\sigma_j^h+k)\big)\,\sum_{i}r_{ij} 
\end{eqnarray}
\noindent
which $k$ is a constant, and $\sum_{i}r_{ij}$ also denotes the amount of data being assigned to $j$. To show, how informations flows through the network, we can compute the following,
\begin{eqnarray}
 a_i = \sum_{j} r_{ij}
\end{eqnarray}
\noindent
Then, in order to determine whether the capsule $j$ will be activated or not, we compute the logistic function of the cost,
\begin{eqnarray}
a_{j} &=& \text{sigmoid} \big( \lambda(b_j- \sum_{h}\text{cost}_{j}^h  )   \big)  \\
&=& \text{sigmoid}\bigg(\lambda(\beta_a- \beta_u\sum_{i}r_{ij}- \text{cost}_j ) \bigg) \nonumber \\
\text{Where:} \quad \text{cost}_j &=& \sum_{i}r_{ij} \text{cost}_{j|i}   \nonumber \\
  &=& \sum_{h}\big(\text{ln}(\sigma_j^h)+\frac{1+\text{ln} 2\pi}{2}\big) \sum_{i}r_{ij}                   \nonumber
\end{eqnarray}
\noindent
where $\beta_a$ denotes the trained parameters for all capsules, and $\beta_u$ denotes the trained parameters for one capsule.
"$-b_j$" (as in the original paper \cite{sabour2018matrix}) is explained as the cost of describing the mean and variance of capsule $j$. In other words, if the benefit $b_j$ of representing datapoints by the parent capsule $j$ outranks the cost caused by the discrepancy in their votes, we activate the output capsules. $b_j$ is not computed analytically. Instead, it should be approximate through training using the backpropagation and a cost function. In the above equations, $r_{ij}$, $\mu$, $\sigma$, and $a_j$ are trained iteratively using EM-routing. $\lambda$ is first initialized to 1 at the beginning of the iterations and then increment by 1 after each routing iteration. It controls the steepness of the sigmoid function. Capsule $j$ is thereby activated when $\sigma_j$ and $\sum_{i}r_{ij}$ are small enough and big enough, respectively.
 The objective of the EM routing is to group capsules to form a part-whole relationship using EM clustering technique. Each of the parts capsules in the lower layer makes predictions (votes) on the pose matrices of its possible parent capsules. Each vote is a predicted value for a parent-capsule's pose matrix. If the parts-related capsules all vote a similar pose matrix value, we cluster them together to form a parent capsule (e.g., in face detection application, if the nose, mouth and eyes capsules all vote a similar pose matrix value, we cluster them together to form a parent capsule: the face capsule). The entire algorithm as in \cite{sabour2018matrix} is depicted, as \vspace{-4mm}
 
\begin{figure}[!h]
	\centering
	\includegraphics[width=1.02\linewidth, height=.3\textheight]{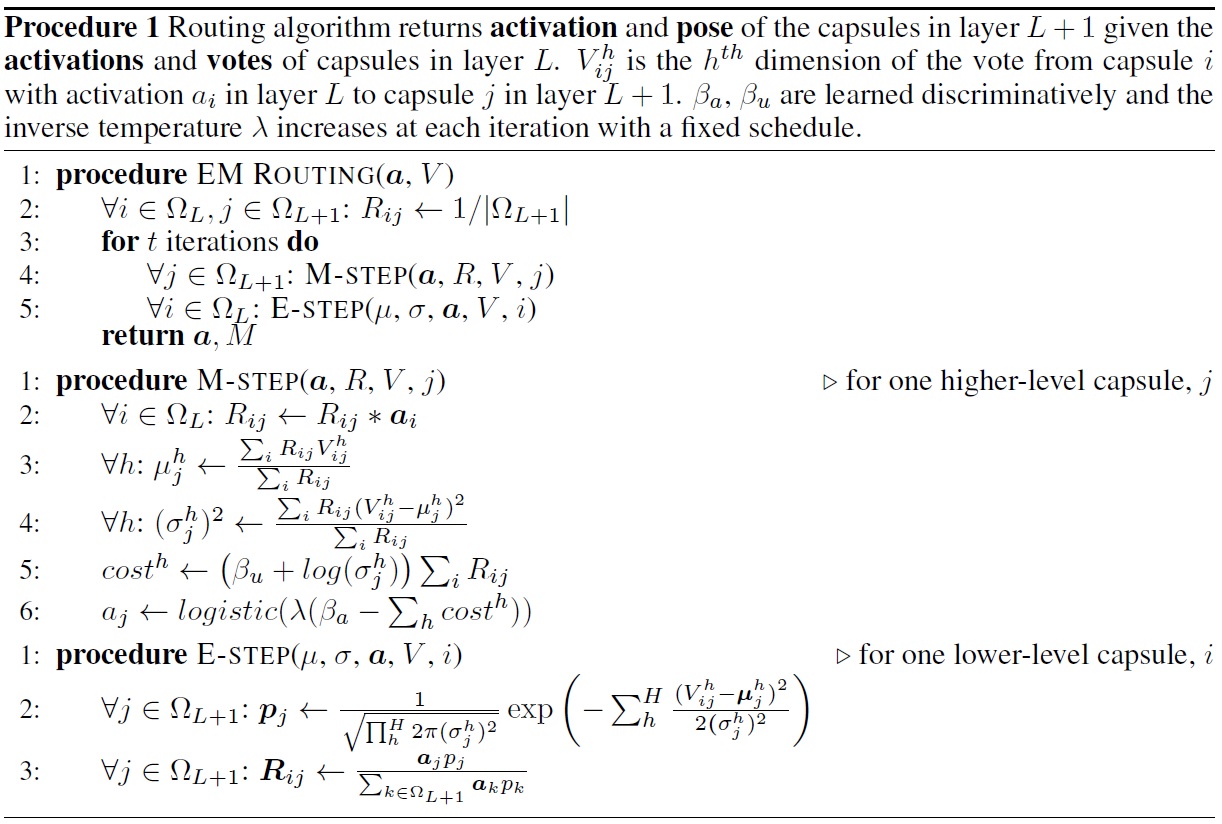}
	\label{fig:EM}
	\vspace{-6mm}
\end{figure}
\noindent
The E-step determines the assignment probability $r_{ij}$ of each datapoint to a parent capsule. The M-step re-calculate the Gaussian models’ values based on $r_{ij}$. In E-step, we re-calculate the assignment probability $r_{ij}$ based on the new $\mu$, $\sigma$ and $a_j$. The assignment is increased if the vote is closer to the $\mu$ of the updated Gaussian model.

In M-step, we calculate $\mu$\ and $\sigma$ based on the activation $a_i$  from the children capsules, the current $r_{ij}$ and votes $\boldsymbol{V}$. M-step also re-calculates the cost and the activation $a_j$  for the parent capsules.

Finally, the “\textit{spread loss}”, $L$, is used to maximize the gap between the activation of the target class ($a_t$) and
the activation of the other classes, considering a margin, $m$, so that if the squared distance between them is smaller than $m$, the loss of this pair is set to zero:
\begin{eqnarray}
L_i = \text{max}\big(0, m-(a_t-a_i)^2  \big) ,\qquad L=\sum_{i\neq t} L_i
\end{eqnarray}

\section{Experimental Study}
The data in our study is provided from \textbf{Hoda} dataset which contains handwritten Persian/Arabic images, which contains $60000$ training-data, as well as $20000$ test-data samples, in grayscale with $32 \times 32$ bits resolution. For cross-validation, $5000$ data samples is considered for each epoch.

The hardware used for the experiments, was a laptop with CPU 4700MQ, Core i7- 2.4GHz, with $8$ GB RAM. All programs are implemented on Tensorflow ver 1.11.0, and Keras ver 2.1.5 software platforms. The training time cost, using the above-mentioned hardware, for every epoch was taken about $75 \sim 80$ minutes.

\begin{figure}[!h]
\centering
\includegraphics[width=1\linewidth, height=.23\textheight]{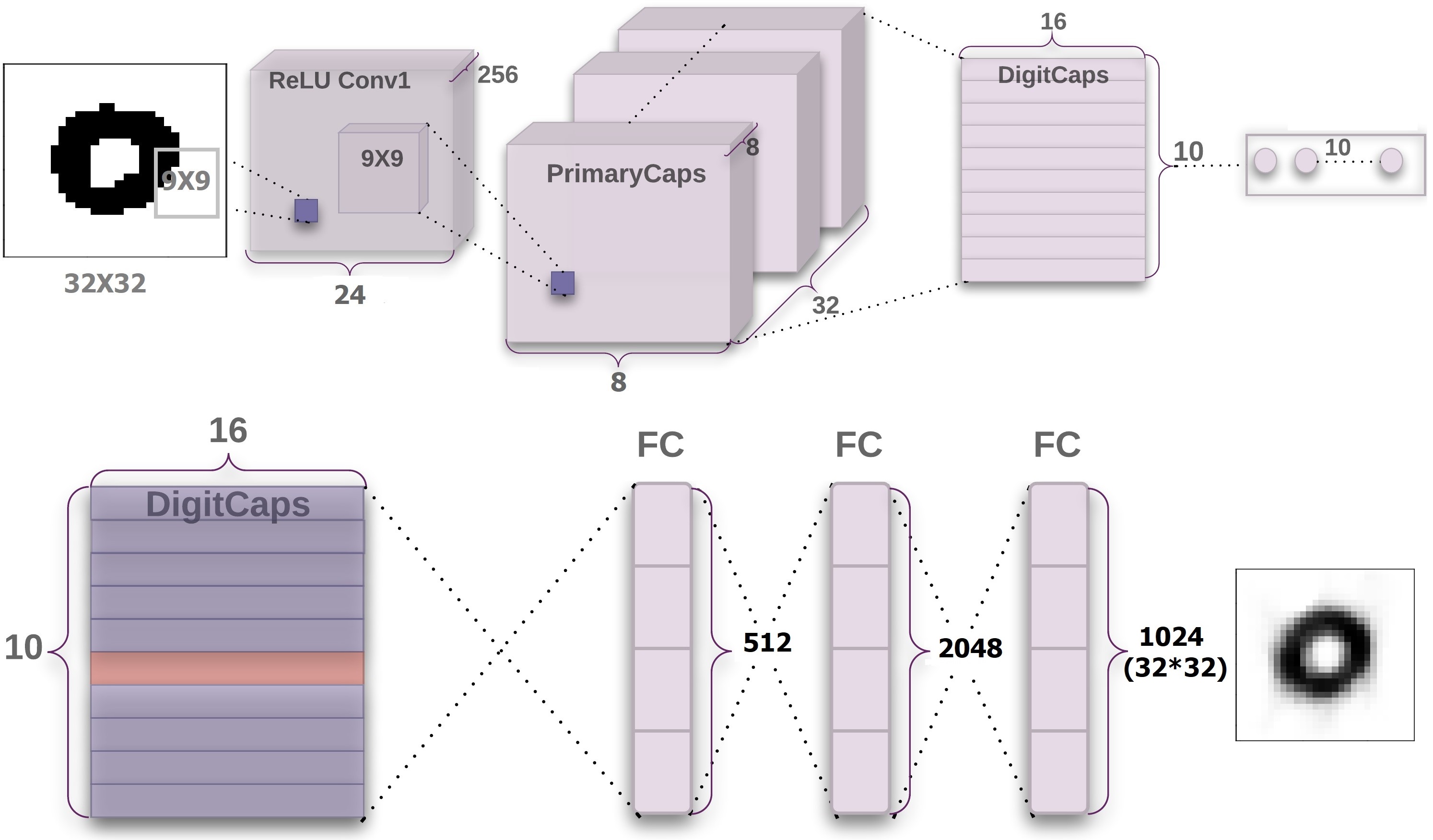}
\caption{Capsule Network digit recognizer and decoder}
\label{fig:caps_scheme_both}
\end{figure}

\begin{figure}[!b]
	\subfloat[\label{subfig-1:tr}]{%
		\includegraphics[width=0.44\textwidth,height=.09\textheight]{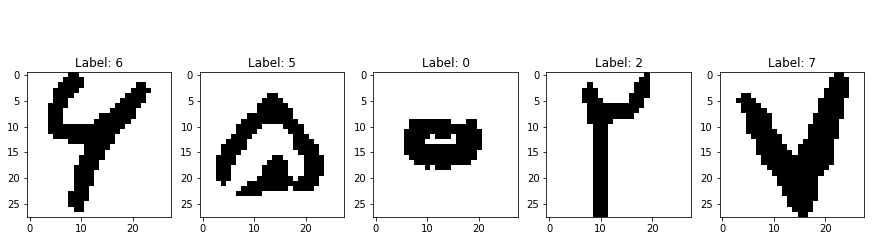}
	}
	\hfill
	\subfloat[\label{subfig-2:ts}]{%
		\includegraphics[width=0.44\textwidth,height=.08\textheight]{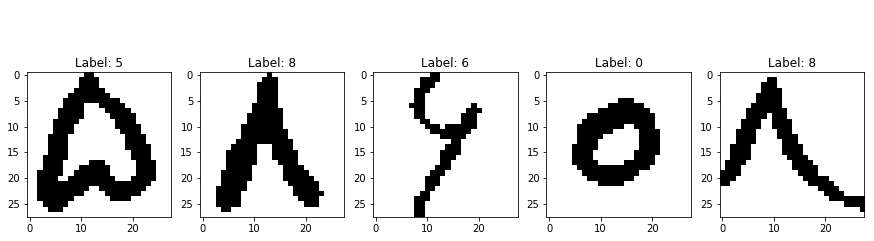}
	}
	\caption{A sample data taken from Hoda dataset; (a) A Training sample, (b) A Test sample }
	\label{fig:DataTrTs}
\end{figure}

\begin{figure}[!h]
	\subfloat[\label{subfig-1:tSNE1}]{%
		\includegraphics[width=0.24\textwidth]{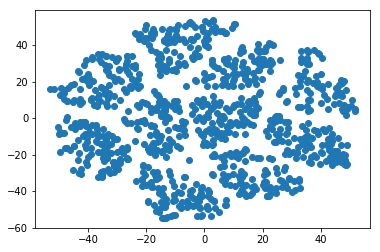}
	}
	\subfloat[\label{subfig-2:tSNE2}]{%
		\includegraphics[width=0.24\textwidth]{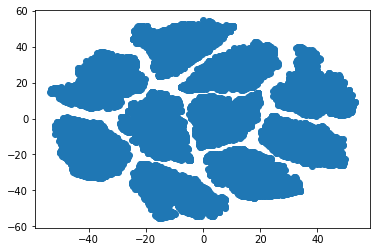}
	}
	\caption{tSNE-based Clustering of Persian/Arabic handwritten  digits using samples from the Hoda dataset, (a) 1000 samples, (b) 20000 samples}
	\label{fig:tSNE}
\end{figure}

\begin{figure}[!h]
\centering
\includegraphics[width=0.95\linewidth, height=.19\textheight]{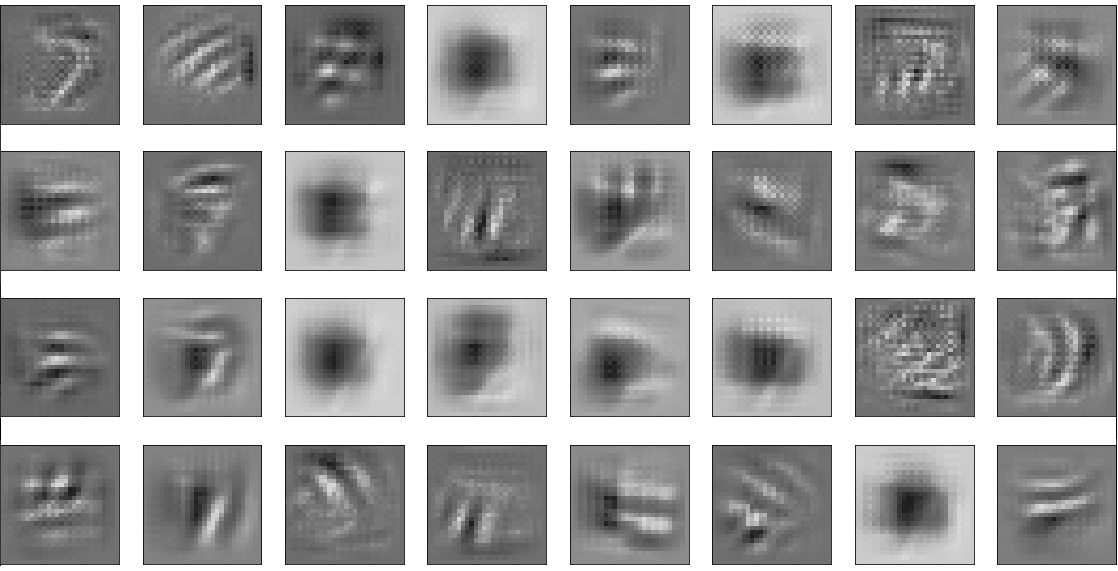}
\caption{Representation of the capsule data; $8$ capsules in each row, versus $4$ epochs in each column}
\label{fig:capsuls_representations}
\end{figure}

\begin{figure}[!h]
	\centering
	\subfloat[\label{subfig-1:decoderinit}]{
	\includegraphics[width=0.9\linewidth, height=.1\textheight]{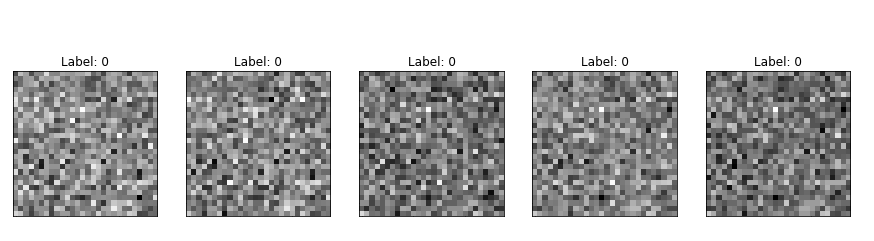}
	}
	\\
	\subfloat[\label{subfig-2:decodEpoch1}]{%
		\includegraphics[width=0.46\textwidth,height=.09\textheight]{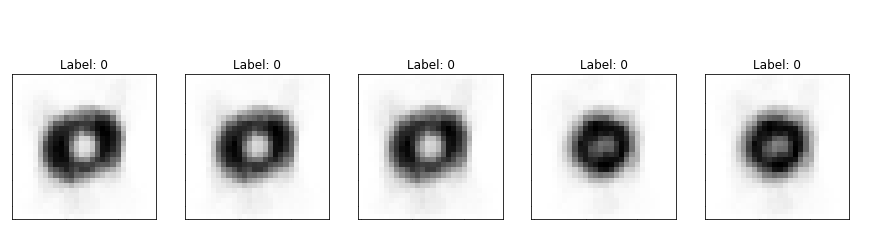}
	}
	\\
	\subfloat[\label{subfig-3:decodEpoch2}]{%
		\includegraphics[width=0.45\textwidth,height=.09\textheight]{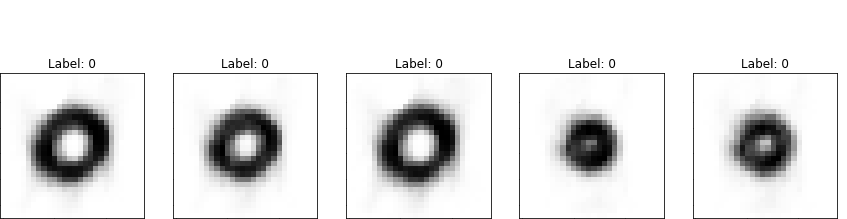}
	}
	\caption{CapsNet decoder outputs after: (a) initialization (b) epoch-1 (c) epoch-2}
	\label{fig:decoderOutput}
\end{figure}

The high-level architecture of the CapsNet used in this paper is depicted in figure~\ref{fig:caps_scheme_both}.
\noindent
A sample data used for training our CapsNet, and for testing it are depicted in figure\ref{fig:DataTrTs}.
The spread of training data in our dataset has been evaluated using \textit{tSNE} algorithm \cite{maaten2008visualizing,van2014accelerating}, with two different 
numbers of data samples of 1000 and 20000, respectively. These clusterings are depicted in figure~\ref{fig:tSNE}.
A representation of the capsule data is extracted which is depicted in figure\ref{fig:capsuls_representations}. 
Moreover, a generated image set from the CapsNet decoder as the initial value, the output of the decoder after epoch 1, and after epoch 2, for digit zero, are respectively depicted in figure~\ref{fig:decoderOutput}.

\begin{figure}
\centering
\includegraphics[width=1.01\linewidth, height=.32\textheight]{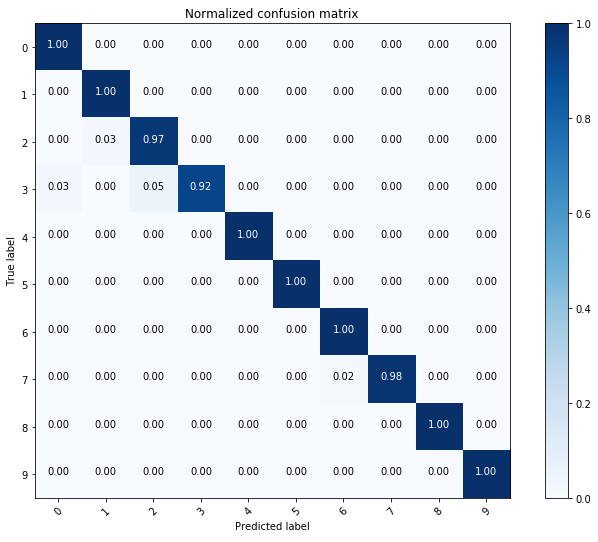}
\caption{Normalized confusion matrix for Persian/Arabic handwritten digits of \textbf{Hoda} dataset, using the proposed CapsNet}
\label{fig:normal_confusion_matrix}
\end{figure}

Now, by feeding the test data in the trained network with the explained architecture, the normalized confusion matrix as in figure~\ref{fig:normal_confusion_matrix} was achieved.

\begin{figure}[!t]
\centering
\includegraphics[width=1\linewidth, height=.55\textheight]{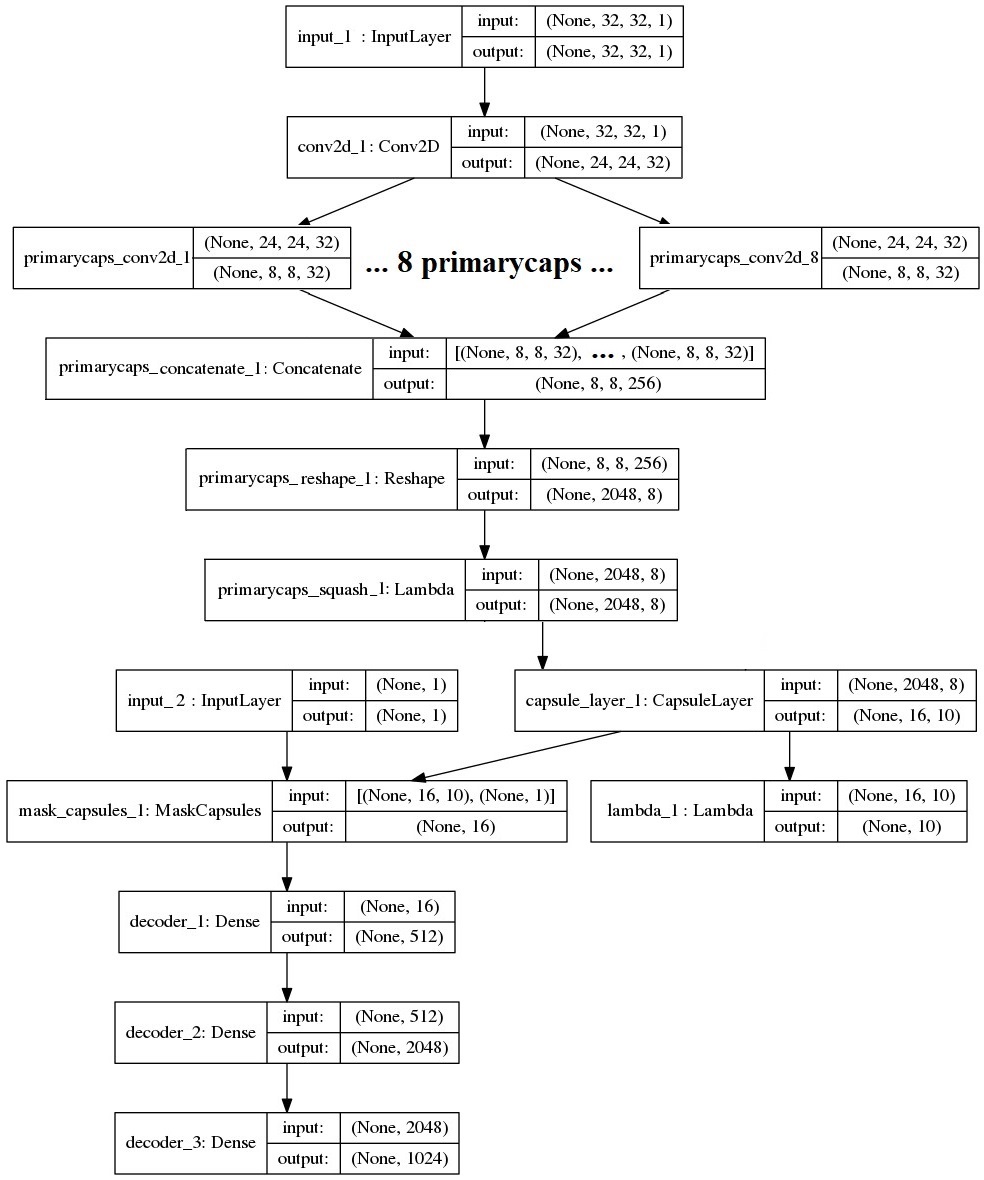}
\caption{Detailed Tree-graph of the capsule network architecture, including the input layer, convolution layers, primary capsules, decoders, and associated parameter settings.}
\label{fig:arch1}
\end{figure}

 The complete architecture of the capsule network and detailed parameters are depicted in figure \ref{fig:arch1}.
The confusion matrix of figure~\ref{fig:normal_confusion_matrix}, is obtained using a 400-samples batch size. However, for most of the confusion matrices, the recognizer was $100\%$ successful in classifying the digits. For those otherwise, this figure depicts the worst case happening in the recognition task.

As it was primarily expected, the capsule architecture has equivariance property, which makes it inherently robust against the rotation, transformation and different viewpoints. This property has been properly leveraged in this work to recognize the handwritten digits which are gathered from human in a real environment. As it was already proven to be successful in MNIST recognition task, here it is emphasized that the CapsNet architecture could be involved as the optimum recognizer, compared to the state-of-the-art.

\begin{table}
	\centering
	 \caption{Comparison of algorithms on Hoda dataset }
\begin{tabular}{|c|c|c|c|c|}
	\hline 
	\multirow{2}{*}{Algorithms} & \multicolumn{2}{c|}{Dataset Size} & \multicolumn{2}{c|}{Accuracy(\%)}\tabularnewline
	\cline{2-5} \cline{3-5} \cline{4-5} \cline{5-5} 
	& Train & Test & Train & Test\tabularnewline
	\hline 
	\hline 
	Sadri et al. \cite{sadri2003application} & 7390 & 3035 & --- & 94.14\tabularnewline
	\hline 
	Mozaffari et al. \cite{mozaffari2005structural} & 2240 & 1600 & 100 & 94.44\tabularnewline
	\hline 
	Alaei et al. \cite{alaei2009using} & 60000 & 20000 & 99.99 & 98.71\tabularnewline
	\hline 
	Alei et al. \cite{alaei2009fine} & 60000 & 20000 & 99.99 & 99.02\tabularnewline
	\hline 
	\textbf{Proposed} & 60000 & 20000 & 100 & \textbf{99.87}\tabularnewline
	\hline 
	\end{tabular}
	\label{tbl:comparison}	
\end{table}

\begin{figure}[!t]
	\centering
	\includegraphics[width=1\linewidth, height=.09\textheight]{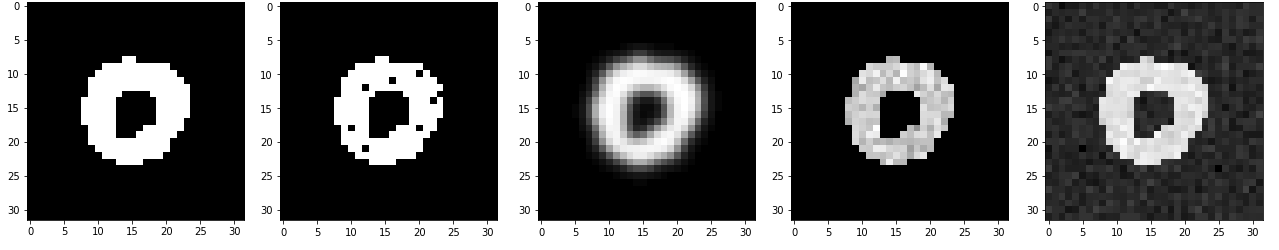}
	\caption{Noisy images being augmented to the input data, also could not fool the classifier;(left to right): input image, pepper noise, Blurred image, Speckle noise, Gaussian Noise}
	\label{fig:noise}
\end{figure}

The results of the proposed technique are further compared with some of the state-of-the-art methods which have been already applied on the same dataset, and are shown in Table~\ref{tbl:comparison}. Similar to Alaei et al. \cite{alaei2009fine}, we used $60000$ samples for training our CapsNet, and $20000$ samples out of Hoda dataset for testing the system. As it is clearly obvious through Table~\ref{tbl:comparison}, the results of our proposed system outperforms the other techniques, even during classification of the noisy input images. For that purpose, we further created some noisy images using different common noises (e.g., pepper noise, Blurring image, Speckle noise, Gaussian noise) and fed them through the classifier, as in Figure~\ref{fig:noise}. Even for the noisy images the system could perfectly classify the input digits. 

The total number of misclassified samples out of $20000$ was about 26 samples. The total number of trainable parameters are $6,455,376$. These parameters are gathered in a model with almost $98.5$ MBytes capacity.

An advantage of the proposed system over the methods shown in Table~\ref{tbl:comparison}, is the robustness of it against the noise. All the other successful methods mentioned in Table~\ref{tbl:comparison}, use SVM classifier in the core of the algorithm. It is known from theory, that SVM classifier is highly sensible against the noise and outliers. Moreover, the Capsule network has the equivariance property which makes it robust against various transformations over the input images. Most of the conventional classifiers require pre-processing blocks to get rid of the initial transformations and then they are applied to the inputs. This inherent ability of CapsNets, justifies its usage despite the large model parameters it can impose to the model training.

\noindent

\section{Conclusion}
In this paper the problem of handwritten digit recognition for Persian/Arabic language, was addressed. For the first time, capsule network was employed and trained using EM algorithm. By comparing the results achieved on Hoda dataset, with the state-of-the-art on the same dataset, we could see the superiority of the capsule network on recognizing the handwritten digits, using all the previously applied methods. The power of the system against the noisy input images are further investigated, and the outputs have been proven to be robust against various noise effects and transformations.

\section*{Acknowledgment}
Authors of this paper express special thanks to \textit{Huadong Liao} for providing the Capslayer tools for faster implementation of the capsule network. Further, we acknowledge \textit{Amir Sanayian} for providing the Hoda dataset loader.
\vspace{4mm}

\bibliographystyle{IEEEtran}
\bibliography{Ipria_ref}

\end{document}